\documentclass{article}

\usepackage{amsthm}
\usepackage{amsmath}

\usepackage[final]{nips_2018}

\usepackage[utf8]{inputenc} 
\usepackage[T1]{fontenc}    
\usepackage{hyperref}       
\usepackage{url}            
\usepackage{algpseudocode}
\usepackage{algorithm}
\usepackage{booktabs}       
\usepackage{amsfonts}       
\usepackage{nicefrac}       
\usepackage{microtype}      
\usepackage{mathtools}
\usepackage[lofdepth,lotdepth]{subfig}
\usepackage{wrapfig}
\usepackage{pdfpages}
\usepackage{enumitem}
\usepackage[title]{appendix}
\usepackage{mathtools}

\title{On the Implicit Assumptions of GANs}

\author{
Ke Li \qquad Jitendra Malik \\
Department of Electrical Engineering and Computer Sciences\\
University of California, Berkeley\\
Berkeley, CA 94720\\
United States\\
\texttt{\{ke.li,malik\}@eecs.berkeley.edu}
}

\begin{document}

\maketitle

\begin{abstract}
Generative adversarial nets (GANs)~\citep{goodfellow2014generative,gutmann2014likelihood} have generated a lot of excitement. Despite their popularity, they exhibit a number of well-documented issues in practice, which apparently contradict theoretical guarantees. A number of enlightening papers, e.g.:~\citep{arora2017generalization,sinn2017non,cornish2018towards}, have pointed out that these issues arise from unjustified assumptions that are commonly made, but the message seems to have been lost amid the optimism of recent years. We believe the identified problems deserve more attention, and highlight the implications on both the properties of GANs and the trajectory of research on probabilistic models. We recently proposed an alternative method~\citep{li2018implicit} that sidesteps these problems. 
\end{abstract}

\section{Introduction}

Generative adversarial nets (GANs)~\citep{goodfellow2014generative,gutmann2014likelihood} are one of the most popular generative models today. In the ensuing discussion, we separate the model, also known as the generator, from the training objective for clarity. The model is an example of an implicit probabilistic model~\citep{diggle1984monte,mohamed2016learning}, which is a probabilistic model that is defined most naturally in terms of a sampling procedure. In the case of GANs, the sampling procedure associated with the model is the following:
\begin{enumerate}
\item Sample $\mathbf{z} \sim \mathcal{N}(0,\mathbf{I})$
\item Return $\mathbf{x} \coloneqq T_{\theta}(\mathbf{z})$
\end{enumerate}
where $T_{\theta}(\cdot)$ is a neural net.

For implicit models, marginal likelihood in general cannot be expressed analytically or computed numerically. Consequently, implicit models cannot be trained by directly maximizing likelihood and must instead be trained using likelihood-free approaches, which do not require evaluation of the likelihood or any derived quantities. GANs use adversarial loss as their training objective, which penalizes dissimilarity of samples to data. It has been shown that when given access to an infinitely powerful discriminator, the original GAN objective minimizes the Jensen-Shannon divergence, the $-\log D$ variant of the objective minimizes the reverse KL-divergence minus a bounded quantity~\citep{arjovsky2017towards}, and later extensions~\citep{nowozin2016f} minimize arbitrary \emph{f}-divergences. Various papers~\citep{arora2017generalization,sinn2017non,cornish2018towards} have pointed out that these results only hold under strong assumptions, but these are often forgotten. These assumptions have far-reaching implications on both the properties of adversarial loss and the trajectory of research on probabilistic models, and so we believe these deserve more attention than they have received. In this paper, we highlight the most important issues, discuss their implications and propose an alternative method that bypasses these issues. 

\section{True vs. Empirical Data Distribution}

As pointed out by~\citep{arora2017generalization,sinn2017non,cornish2018towards}, theoretical analysis of GANs often disregards the distinction between the true and the empirical data distribution. More concretely, the original adversarial training objective is:
\[
\min_{\theta_{G}}\max_{\theta_{D}}\mathbb{E}_{\mathbf{x}\sim p_{\mathrm{data}}}\left[\log D_{\theta_{D}}(\mathbf{x})\right]+\mathbb{E}_{\mathbf{z}\sim p_{z}}\left[\log\left(1-D_{\theta_{D}}\left(G_{\theta_{D}}(\mathbf{z})\right)\right)\right]
\]
where $p_{data}$ is the \emph{true} data distribution. Ideally, both expectations should be approximated with Monte Carlo estimates during training to ensure unbiased gradient estimates. However, because drawing samples from the true data distribution is expensive and usually cannot be done during training, $\mathbb{E}_{\mathbf{x}\sim p_{\mathrm{data}}}\left[\log D_{\theta_{D}}(\mathbf{x})\right]$ is in practice approximated with samples drawn from the training dataset rather than from the true data distribution. Drawing samples from the training dataset effectively results in a Monte Carlo estimate of the expectation w.r.t. the \emph{empirical} data distribution. While this is an unbiased estimate of the expectation w.r.t. the \emph{empirical} data distribution, it is not necessarily an unbiased estimate of the expectation w.r.t. the \emph{true} data distribution, because the randomness in the data collection process is not marginalized out when training the model.~\footnote{More details are in Appendix~\ref{sec:expectations}.}

While seemingly innocuous, the replacement of the true data distribution with the empirical version has important consequences. For example, \citet{arora2017generalization} pointed out that because the empirical data distribution is discrete, the Jensen-Shannon divergence between the empirical data distribution and \emph{any} continuous model distribution is always $\log 2$, which is the maximum possible value of the Jensen-Shannon divergence. Therefore, minimizing the Jensen-Shannon divergence between the empirical data distribution and the model distribution does not necessarily recover the true data distribution. 

Minimizing reverse KL-divergence, i.e. the KL-divergence from the empirical data distribution to the model distribution, $D_{KL}(p_{\theta}\Vert\widehat{p_{\mathrm{data}}})$, is similarly problematic. Recall that $D_{KL}(p \Vert q)$ is only defined and finite if the $p$ is absolutely continuous w.r.t. $q$, i.e. for any $x$ such that $q(x) = 0$, $p(x) = 0$. Therefore, for reverse KL-divergence to be defined and finite, the support of the model distribution must be contained in the support of the empirical data distribution, which is just the set of training examples. Therefore, any model that can produce a novel sample that is not in the training set will give rise to an undefined or infinite reverse KL-divergence. In particular, the reverse KL-divergence for any continuous model distribution is undefined, and so minimizing reverse KL-divergence does not make sense.~\footnote{This argument conveys the right intuition, but there are some technical subtleties involved due to the non-existence of the density of the empirical data distribution. A more rigorous treatment can be found in Appendix~\ref{sec:kl}.}

Therefore, claims about GANs minimizing Jensen-Shannon divergence or the reverse KL-divergence minus a bounded quantity implicitly carry the assumption of access to the true data distribution during training. Because this is a strong assumption that rarely holds in practice, it is important to keep this assumption in mind when interpreting theoretical results and applying GANs in practice. 

\section{Asymptotic Consistency}

It is often claimed that the parameter estimate obtained by optimizing the original adversarial objective when the discriminator has infinite capacity (which corresponds to minimizing the Jensen-Shannon divergence)~\citep{goodfellow2014distinguishability} or by minimizing the reverse KL-divergence~\citep{huszar2015not} is \emph{asymptotically consistent}. The notion of asymptotic consistency used in the sense here refers to the fact that when the \emph{true} data distribution is given and the model has sufficient capacity to reproduce the true data distribution, the objective attains the unique global optimum when the model distribution coincides with the true data distribution. It is important to note, however, that this notion of asymptotic consistency is \emph{different} from the classical notion of asymptotic consistency in statistics, and the two should not be conflated. In fact, minimizing \emph{any} divergence from or to the \emph{true} data distribution is asymptotically consistent in the sense as used in the GAN literature, since by definition of divergences, they must always be non-negative and can only evaluate to zero if and only if the two distributions passed in as arguments are equal, implying that the optimum must occur when the model distribution is equal to the true data distribution. 

Unfortunately, the notion of asymptotic consistency used in the GAN literature does not characterize the behaviour of the optima of various objectives in practice, because the true data distribution is almost never available. (If it were available, there would be no need to train a generative model, since we can simply use the true data distribution as given in place of the generative model.) What is more useful is to look at how the parameter estimate behaves when given a \emph{finite} set of training examples, and examine what this parameter estimate converges to as the number of training examples tends to infinity. This is what the classical notion of asymptotic consistency in statistics refers to. 

More precisely, given an infinite sequence of i.i.d. samples $\{\mathbf{x}_i\}_{i=1}^{\infty}$ from $p_{\theta}$, where the true parameter $\theta$ is unknown, let $\hat{\theta}_n$ be a parameter estimator that depends on only the first $n$ samples $\{\mathbf{x}_i\}_{i=1}^{n}$. An estimator $\hat{\theta}_n$ is (weakly) asymptotically consistent if $\mathrm{plim}_{n\rightarrow\infty} \hat{\theta}_n(\{\mathbf{x}_i\}_{i=1}^{n}) = \theta$, where the probability is over the randomness in the process of drawing samples $\{\mathbf{x}_i\}_{i=1}^{\infty}$ from $p_{\theta}$. 

Let's now compare this notion of asymptotic consistency to the notion of asymptotic consistency used in the GAN literature by way of examples. Recall that the Jensen-Shannon divergence between the empirical data distribution and any continuous model distribution is $\log 2$, regardless of what the model distribution is or how many samples there are. Therefore, \emph{any} parameter value minimizes the Jensen-Shannon divergence, because it is a constant function in the parameter value. Therefore, the parameter estimate obtained by minimizing the Jensen-Shannon divergence is not asymptotically consistent in the classical statistical sense. Similarly, because the reverse KL-divergence is undefined for any continuous model distribution, the parameter value that minimizes reverse KL-divergence is undefined, and so the parameter estimate obtained by minimizing reverse KL-divergence is not asymptotically consistent either. 

Therefore, even though minimizing \emph{any} divergence is consistent according to the notion of asymptotic consistency used in the GAN literature, this is \emph{not} true if the notion of asymptotic consistency considered were to agree with the definition in the statistical literature. One divergence that \emph{is} asymptotically consistent in both the statistical sense and the GAN sense is the standard KL-divergence, i.e. the KL-divergence from the model distribution to the empirical data distribution, $D_{KL}(\widehat{p_{\mathrm{data}}} \Vert p_{\theta})$, which is equivalent to maximum likelihood. 

\section{Mode Dropping and Its Implications}

Mode dropping is a well-documented issue in GANs and refers to the phenomenon where the model disregards some modes in the data distribution and assigns them low density. Various theoretical explanations have been proposed~\citep{arora2017generalization,arjovsky2017towards}, which suggest that it could be caused by a combination of factors, including the polynomial capacity of the discriminator and the minimization of reverse KL-divergence. More specifically, \citet{arora2017generalization} showed that a discriminator with a polynomial number of parameters in the dimensionality of the data cannot in general detect mode dropping, leading to the generator at convergence dropping an exponential number of modes. \citet{arjovsky2017towards} demonstrated that even if we assume access to the true data distribution and a discriminator with infinite capacity, the $-\log D$ variant of the adversarial objective, which is more commonly used than the original objective due to issues with vanishing gradients, minimizes the reverse KL-divergence minus a bounded quantity. Because reverse KL-divergence heavily penalizes the model assigning high density to unlikely data and only mildly penalizes the model assigning low density to likely data, it tends to lead to a model distribution that misses some modes of the data distribution. 

The freedom that GANs have to drop modes has important implications on the evaluation of generative models and the trajectory of research on generative models more broadly. It is instructive to think of the performance of a generative model along two axes: precision, i.e.: its ability to generate plausible samples, and recall, i.e.: its ability to model the full diversity of the data distribution. Ideally, we would like to learn a model that scores high along both axes. Traditionally, generative models were trained using maximum likelihood. Because likelihood is the product of densities at each of the training examples, such models are not allowed to assign low density to any of the training examples, because overall likelihood would become low. So, mode dropping is effectively disallowed and full recall is guaranteed. Evaluation in this case is straightforward -- since recall is fixed, an increase in precision implies an upward shift in the precision-recall curve, which indicates better modelling performance. So, models trained using maximum likelihood can be compared solely on the basis of precision. Precision can be easily measured -- a simple way is by visual assessment of sample quality. This is why sample quality has historically been an important indicator of modelling performance and why sample quality used to correlate with log-likelihood (or estimates/lower bounds thereof). 

On the other hand, GANs are allowed to drop modes and can effectively choose the data examples that it wants to model. Because the model designer has no control over which or how many training examples are ignored, there is no guarantee on the level of recall, and so it is critical to measure both precision and recall. To see why both precision and recall are important, consider a model with low capacity that drops all but a few modes. Such a model is able to trivially achieve high precision by dedicating all its modelling capacity to the few modes, but cannot explain most of the data. Evaluation in this case becomes tricky -- an increase in precision could mean either a movement along the precision-recall curve, where the increase in precision comes at the expense of a decrease in recall, or an upward shift in the precision-recall curve, where recall is maintained or increased together with the improvement in precision. Only the latter implies an improvement in modelling performance. As a result, evaluation by precision no longer suffices and can be misleading. Instead, both precision and recall must be measured and care must be taken to interpret the results. 

Unfortunately, to date, there has been no reliable method to measure recall. Visualization of samples no longer works because human memory constraints limit our ability to detect a deficiency in diversity compared to the training data. Comparison to the nearest training example is also problematic because it only detects almost exact memorization of a training example. As a result, recall is typically not measured. This is dangerous -- without measuring recall, we do not know how much recall was given up in order to achieve an improvement in sample quality. 

This has had an impact on the trajectory of research in probabilistic modelling, which has deviated somewhat from the original goal, which is to model the inherent uncertainty in inference/prediction. Performance of generative models was traditionally measured in terms of log-likelihood or a lower bound on log-likelihood, and assessment of sample quality was a way of visualizing performance. Later on, estimated log-likelihoods replaced lower bounds on log-likelihoods as a performance metric, since there was no straightforward way to compute lower bounds on log-likelihoods for some models. Fortunately, sample quality was unaffected by this limitation and served to validate conclusions drawn from estimated log-likelihoods, which may be unreliable in high dimensions. The advent of models that can drop modes brought about a shift from the full-recall setting to a setting where a level of recall is not guaranteed. Consequently, conclusions drawn from estimated log-likelihood and sample quality diverged. Due to the unreliability of estimated log-likelihood, it has gradually fallen out of favour as a performance metric, leaving sample quality as the sole performance metric. However, it is sometimes forgotten that because full recall is no longer enforced, sample quality no longer reflects how well the model learns the underlying distribution. As a result, research in the area has undergone a largely unnoticed change in focus, from trying to learn the distribution to trying to synthesize visually pleasing images. While the latter is a worthy goal, the former should not be abandoned and researchers should be mindful of the fact that impressive advances in sample quality in recent years do \emph{not} mean that we are now within reach of being able to model the distribution of natural images. We can only claim to have achieved the latter when we can produce realistic samples \emph{at full recall}, which could be considerably more challenging than producing realistic samples at some unknown level of recall. 

\section{Proposed Solution: Implicit Maximum Likelihood Estimation}

How can we as a community progress towards generative models that are capable of learning the underlying data distribution? We argue that the first step is to return to the principle of maximum likelihood and insist on full recall, for the following reason: when the model is given the freedom to drop modes and effectively ignore training data that it does not want to model, it is difficult to design high-capacity models. Even when a model is trained on a large and diverse dataset, model may appear to overfit to a small subset of training examples and fail to generalize, because a lot of examples in the dataset are effectively ignored. The natural course of action to mitigate this would be to reduce the capacity of the model, however, this would result in a low-capacity model that captures only the modes represented by a small subset of the training examples. 

This does not mean that we would have to give up implicit probabilistic models, which do offer a lot more modelling flexibility compared to classical probabilistic models. Previously, it was unclear how to train implicit models to maximize likelihood when the likelihood and derived quantities can neither be expressed analytically or computed numerically. Recently, we introduced a simple likelihood-free parameter estimation technique that can be shown to be equivalent to maximum likelihood under some conditions, which we call Implicit Maximum Likelihood Estimation~\citep{li2018implicit}. 

\footnotesize
\bibliography{gan_assumptions}
\bibliographystyle{icml2018}

\newpage

\begin{appendices}

\section{Monte Carlo Estimates of Expectations}
\label{sec:expectations}

We first show that a Monte Carlo estimate based on samples from the training data is an unbiased estimate of the expectation w.r.t. the \emph{empirical} data distribution, i.e.: $\mathbb{E}_{\mathbf{x}\sim\widehat{p_{\mathrm{data}}}}\left[f(\mathbf{x})\right]$.

Let $\mathbf{x}_{1},\ldots,\mathbf{x}_{n}$ denote the training examples and let $c$ be a categorical distribution with $n$ categories and uniform probabilities over all categories, i.e. if $u \sim c$, $\mathrm{Pr}\left(u = j\right) = 1/n \;\forall j \in \{1,\ldots,n\}$. Then $\frac{1}{m}\sum_{i=1}^{m}\sum_{j=1}^{n}\mathbf{1}\left[u_{i}=j\right]f(\mathbf{x}_{j})$ is the Monte Carlo estimate based on samples from the training data. Note that the expectation is conditioned on $\mathbf{x}_{1},\ldots,\mathbf{x}_{n}$, because training examples are only drawn once while collecting data, and new samples from the true data distribution are not drawn during training. 

\begin{align*}
& \mathbb{E}_{u_{1},\ldots,u_{m}\sim c}\left[\left.\frac{1}{m}\sum_{i=1}^{m}\sum_{j=1}^{n}\mathbf{1}\left[u_{i}=j\right]f(\mathbf{x}_{j})\right|\mathbf{x}_{1},\ldots,\mathbf{x}_{n}\right]\\
= & \frac{1}{m}\sum_{i=1}^{m}\mathbb{E}_{u_{i}\sim c}\left[\left.\sum_{j=1}^{n}\mathbf{1}\left[u_{i}=j\right]f(\mathbf{x}_{j})\right|\mathbf{x}_{1},\ldots,\mathbf{x}_{n}\right]\\ 
= & \frac{1}{m}\sum_{i=1}^{m}\sum_{j=1}^{n}\mathbb{E}_{u_{i}\sim c}\left[\mathbf{1}\left[u_{i}=j\right]\right]f(\mathbf{x}_{j})\\
= & \frac{1}{m}\sum_{i=1}^{m}\sum_{j=1}^{n}\frac{1}{n}f(\mathbf{x}_{j})\\
= & \frac{1}{n}\sum_{j=1}^{n}f(\mathbf{x}_{j})\\
= & \mathbb{E}_{\mathbf{x}\sim\widehat{p_{\mathrm{data}}}}\left[f(\mathbf{x})\right]
\end{align*}

Note that $\mathbb{E}_{\mathbf{x}\sim\widehat{p_{\mathrm{data}}}}\left[f(\mathbf{x})\right] = \frac{1}{n}\sum_{j=1}^{n}f(\mathbf{x}_{j})$ is a random variable (because $\mathbf{x}_{1},\ldots,\mathbf{x}_{n}$ are random), and so cannot be equal to $\mathbb{E}_{\mathbf{x}\sim p_{\mathrm{data}}}\left[f(\mathbf{x})\right]$, which is a constant. Below we show that we can obtain an unbiased estimate of the expectation w.r.t. the true data distribution, i.e.: $\mathbb{E}_{\mathbf{x}\sim p_{\mathrm{data}}}\left[f(\mathbf{x})\right]$, by taking the expectation over the training examples $\mathbf{x}_{1},\ldots,\mathbf{x}_{n}$, which essentially marginalizes out the randomness during the data collection process and requires drawing fresh samples from the true data distribution during training. 

\begin{align*}
& \mathbb{E}_{\mathbf{x}_{1},\ldots,\mathbf{x}_{n}\sim p_{\mathrm{data}}}\left[\mathbb{E}_{u_{1},\ldots,u_{m}\sim c}\left[\left.\frac{1}{m}\sum_{i=1}^{m}\sum_{j=1}^{n}\mathbf{1}\left[u_{i}=j\right]f(\mathbf{x}_{j})\right|\mathbf{x}_{1},\ldots,\mathbf{x}_{n}\right]\right] \\
= & \mathbb{E}_{\mathbf{x}_{1},\ldots,\mathbf{x}_{n}\sim p_{\mathrm{data}}}\left[\frac{1}{n}\sum_{j=1}^{n}f(\mathbf{x}_{j})\right]\\
= & \frac{1}{n}\sum_{j=1}^{n}\mathbb{E}_{\mathbf{x}_{j}\sim p_{\mathrm{data}}}\left[f(\mathbf{x}_{j})\right]\\
= & \frac{1}{n}\sum_{j=1}^{n}\mathbb{E}_{\mathbf{x}\sim p_{\mathrm{data}}}\left[f(\mathbf{x})\right]\\
= & \mathbb{E}_{\mathbf{x}\sim p_{\mathrm{data}}}\left[f(\mathbf{x})\right]
\end{align*}

\section{KL-Divergence between Discrete and Continuous Distributions}
\label{sec:kl}

KL-divergence is typically defined as:
\begin{align*}
D_{KL}\left(P\Vert Q\right) \coloneqq \mathbb{E}_{\mathbf{x}\sim P}\left[\log\left(\frac{dP}{dQ}(\mathbf{x})\right)\right] = \int_{S}\log\left(\frac{dP}{dQ}\right)dP
\end{align*}
where $P$ and $Q$ denote two probability measures and $\frac{dP}{dQ}$ denotes the Radon-Nikodym derivative of $P$ w.r.t. $Q$. This notion of KL-divergence is well-defined when $P$ and $Q$ are both continuous or are both discrete. However, it is not well-defined when one distribution is discrete and the other is continuous. Therefore, under this definition, neither $D_{KL}( \widehat{p_{\mathrm{data}}} \Vert p_{\theta})$ nor $D_{KL}(p_{\theta}\Vert\widehat{p_{\mathrm{data}}})$ is well-defined. 

However, we will show below that under a slightly more general definition of KL-divergence, $D_{KL}\left(P\Vert Q\right)$ \emph{is} well-defined when $P$ is discrete and $Q$ is continuous, but not the other way around. 

Consider the following notion of KL-divergence:
\begin{align*}
D_{KL}\left(P\Vert Q\right) & \coloneqq \mathbb{E}_{\mathbf{x}\sim P}\left[\log p(\mathbf{x})\right]-\mathbb{E}_{\mathbf{x}\sim P}\left[\log q(\mathbf{x})\right] \\
&= \int_{S}\log\left(\frac{dP}{d\nu_{P}}\right)dP-\int_{\mathbb{R}^{d}}\log\left(\frac{dQ}{d\nu_{Q}}(\mathbf{x})\right)dF_{P}(\mathbf{x})
\end{align*}
where $p$ and $q$ denote probability mass functions (PMFs) or probability density functions (PDFs) of $P$ and $Q$ respectively (depending on whether each is discrete or continuous), $\nu_P$ and $\nu_Q$ denote the reference measures for $P$ and $Q$, and $F_P$ and $F_Q$ denote the cumulative distribution functions (CDFs) of $P$ and $Q$ respectively. If $P$ is continuous, the reference measure $\nu_P$ is the Lebesgue measure $\lambda$ on $\mathbb{R}^{d}$; if it is discrete, $\nu_P$ is the counting measure $\mu$ on a set $D$. We will write out the definition of $p$ and $q$ explicitly as Radon-Nikodym derivatives to emphasize the possibly different reference measures for $P$ and $Q$. Note that the second integral is w.r.t. the Lebesgue-Stieltjes measure associated to the CDF of $P$. It is easy to check that when $P$ and $Q$ are both continuous or are both discrete, this definition is equivalent to the previous definition. 

Now we consider the case where $P$ is discrete and $Q$ is continuous. We first evaluate the first term:
\begin{align*}
\mathbb{E}_{\mathbf{x}\sim P}\left[\log p(\mathbf{x})\right] & = \int_{S}\log\left(\frac{dP}{d\nu_{P}}\right)dP \\
& = \int_{D}\log\left(\frac{dP}{d\mu}\right)dP\\
& = \int_{D}\left(\frac{dP}{d\mu}\right) \log\left(\frac{dP}{d\mu}\right) d\mu\\
& = \sum_{\mathbf{x}\in D}\left(\frac{dP}{d\mu}(\mathbf{x})\right)\log\left(\frac{dP}{d\mu}(\mathbf{x})\right) \\
& = \sum_{\mathbf{x}\in D}p(\mathbf{x}) \log p(\mathbf{x})
\end{align*}

So, the first term is clearly well-defined. We then evaluate the second term:
\begin{align*}
\mathbb{E}_{\mathbf{x}\sim P}\left[\log q(\mathbf{x})\right] & = \int_{\mathbb{R}^{d}}\log\left(\frac{dQ}{d\nu_{Q}}(\mathbf{x})\right)dF_{P}(\mathbf{x})\\
& = \int_{\mathbb{R}^{d}}\log\left(\frac{dQ}{d\lambda}(\mathbf{x})\right)dF_{P}(\mathbf{x})\\
& = \sum_{\mathbf{x}\in D} \left(\frac{dP}{d\mu}(\mathbf{x})\right) \log\left(\frac{dQ}{d\lambda}(\mathbf{x})\right)\\
& = \sum_{\mathbf{x}\in D}p(\mathbf{x})\log q(\mathbf{x})
\end{align*}

Since $\log\left(\frac{dQ}{d\lambda}\right)$ is continuous and $F_P$ is of bounded variation, the Lebesgue-Stieltjes integral is equivalent to the Riemann-Stieltjes integral. From the definition of the Riemann-Stieltjes integral, we can see that $\int_{\mathbb{R}^{d}}\log\left(\frac{dQ}{d\lambda}(\mathbf{x})\right)dF_{P}(\mathbf{x})$ evaluates to $\sum_{\mathbf{x}\in D} \left(\frac{dP}{d\mu}(\mathbf{x})\right) \log\left(\frac{dQ}{d\lambda}(\mathbf{x})\right) = \sum_{\mathbf{x}\in D}p(\mathbf{x})\log q(\mathbf{x})$. 

Hence, when $P$ is discrete and $Q$ is continuous, 
\begin{align*}
D_{KL}\left(P\Vert Q\right) & = \sum_{\mathbf{x}\in D}p(\mathbf{x}) \log p(\mathbf{x}) - \sum_{\mathbf{x}\in D}p(\mathbf{x})\log q(\mathbf{x}) = \sum_{\mathbf{x}\in D}p(\mathbf{x})\log\frac{p(\mathbf{x})}{q(\mathbf{x})}
\end{align*}
Note that while this final expression appears similar to the expression for KL-divergence when both $P$ and $Q$ are discrete, $q$ here denotes the PDF rather than the PMF (which doesn't exist because $Q$ is continuous). 

Therefore, under this definition of KL-divergence, $D_{KL}( \widehat{p_{\mathrm{data}}} \Vert p_{\theta}) = \sum_{i=1}^{n}\frac{1}{n}\log\frac{1}{n}-\sum_{i=1}^{n}\frac{1}{n}\log p_{\theta}(\mathbf{x}_{i})$, where $\{\mathbf{x}_i\}_{i=1}^{n}$ are the training examples. This is why minimizing $D_{KL}( \widehat{p_{\mathrm{data}}} \Vert p_{\theta})$ is equivalent to maximizing likelihood. 

On the other hand, when $P$ is continuous and $Q$ is discrete, the first term is:
\begin{align*}
\mathbb{E}_{\mathbf{x}\sim P}\left[\log p(\mathbf{x})\right] & = \int_{S}\log\left(\frac{dP}{d\nu_{P}}\right)dP\\
& = \int_{\mathbb{R}^{d}}\log\left(\frac{dP}{d\lambda}\right)dP\\
&= \int_{\mathbb{R}^{d}}\left(\frac{dP}{d\lambda}\right)\log\left(\frac{dP}{d\lambda}\right)d\lambda\\
&= \int_{\mathbb{R}^{d}}p(\mathbf{x})\log p(\mathbf{x})d\mathbf{x}
\end{align*}
This is just differential entropy and is clearly well-defined. The second term is:

\begin{align*}
\mathbb{E}_{\mathbf{x}\sim P}\left[\log q(\mathbf{x})\right] & = \int_{\mathbb{R}^{d}}\log\left(\frac{dQ}{d\nu_{Q}}(\mathbf{x})\right)dF_{P}(\mathbf{x})\\
& = \int_{\mathbb{R}^{d}}\log\left(\frac{dQ}{d\mu}(\mathbf{x})\right)dF_{P}(\mathbf{x})
\end{align*}

This is not well-defined, because $\frac{dQ}{d\mu}(\mathbf{x})$ is not defined for $\mathbf{x} \notin D$. This implies that under this definition, even though $D_{KL}( \widehat{p_{\mathrm{data}}} \Vert p_{\theta})$ is well-defined, reverse KL-divergence, $D_{KL}( p_{\theta} \Vert \widehat{p_{\mathrm{data}}} )$, is not. 

\end{appendices}

\end{document}